\documentclass[runningheads]{llncs}

\usepackage[mobile]{accv}

\usepackage{accvabbrv}

\usepackage{graphicx}
\usepackage{booktabs}

\usepackage[accsupp]{axessibility}  

\usepackage[pagebackref,breaklinks,colorlinks,citecolor=accvblue]{hyperref}

\usepackage{orcidlink}

\usepackage{url}

\usepackage{amsmath, amsfonts}
\usepackage{soul, color}
\usepackage{graphicx}
\usepackage{wrapfig,lipsum}
\usepackage{booktabs}
\usepackage{tikz}
\usepackage{pgfplots}
\usepgfplotslibrary{fillbetween}
\usepackage{enumitem}
\usepackage{wrapfig,lipsum}

\usepackage{pifont}
\newcommand{\yes}{\ding{51}}%
\newcommand{\no}{\ding{55}}%

\DeclareMathOperator*{\argmax}{arg\,max}
\newcommand{\bb}{\mathbf{b}}
\newcommand{\bbR}{\mathbb{R}}
\newcommand{\cC}{\mathcal{C}}
\newcommand{\bC}{\mathbf{C}}

\newcommand{\cL}{\mathcal{L}}

\newcommand{\cX}{\mathcal{X}}
\newcommand{\bT}{\mathbf{T}}
\newcommand{\bS}{\mathbf{S}}
\newcommand{\bO}{\mathbf{O}}
\newcommand{\bo}{\mathbf{o}}

\newcommand{\cY}{\mathcal{Y}}

\usepackage{array}
\newcolumntype{C}[1]{>{\centering\let\newline\\\arraybackslash\hspace{0pt}}m{#1}}

\begin{document}

\title{Weakly Supervised Test-Time Domain Adaptation for Object Detection} 

\titlerunning{Weakly Supervised Test-Time Domain Adaptation for Object Detection}


\author{Anh-Dzung Doan\inst{1} \and
Bach Long Nguyen\inst{2} \and
Terry Lim\inst{3} \and Madhuka Jayawardhana\inst{3} \and Surabhi Gupta\inst{3} \and Christophe Guettier\inst{3} \and Ian Reid\inst{1} \and Markus Wagner\inst{2} \and Tat-Jun Chin\inst{1}}

\authorrunning{Anh-Dzung Doan et al.}

\institute{Australian Institute for Machine Learning, The University of Adelaide \and Department of Data Science and Artificial Intelligence, Monash University \and Safran}

\maketitle

\begin{abstract}
  Prior to deployment, an object detector is trained on a dataset compiled from a previous data collection campaign. However, the environment in which the object detector is deployed will invariably evolve, particularly in outdoor settings where changes in lighting, weather and seasons will significantly affect the appearance of the scene and target objects. It is almost impossible for all potential scenarios that the object detector may come across to be present in a finite training dataset. This necessitates continuous updates to the object detector to maintain satisfactory performance. Test-time domain adaptation techniques enable machine learning models to self-adapt based on the distributions of the testing data.  However, existing methods mainly focus on fully automated adaptation, which makes sense for applications such as self-driving cars. Despite the prevalence of fully automated approaches, in some applications such as surveillance, there is usually a human operator overseeing the system's operation. We propose to involve the operator in test-time domain adaptation to raise the performance of object detection beyond what is achievable by fully automated adaptation. To reduce manual effort, the proposed method only requires the operator to provide weak labels, which are then used to guide the adaptation process. Furthermore, the proposed method can be performed in a streaming setting, where each online sample is observed only once. We show that the proposed method outperforms existing works, demonstrating a great benefit of human-in-the-loop test-time domain adaptation. Our code is publicly available at  \url{https://github.com/dzungdoan6/WSTTA}
  \keywords{Human in the loop \and Test-time domain adaptation \and Object detection}
\end{abstract}

\section{Introduction}
\label{sec:intro}

Object detection is a task that involves precisely localising and categorising objects within an image. It has many applications in autonomous driving~\cite{han2021soda10m}, surveillance~\cite{lu2023cross}, and augmented reality~\cite{li2020object}. The deployment of an object detector typically includes three main steps. Firstly, a large-scale dataset must be collected and annotated, providing the bounding boxes and object categories. Next, this annotated dataset is used to train an object detector. Finally, the object detector is deployed into a desired system to effectively perform real-time object detection.

\begin{figure*}
    \centering
    \includegraphics[width=1.0\textwidth]{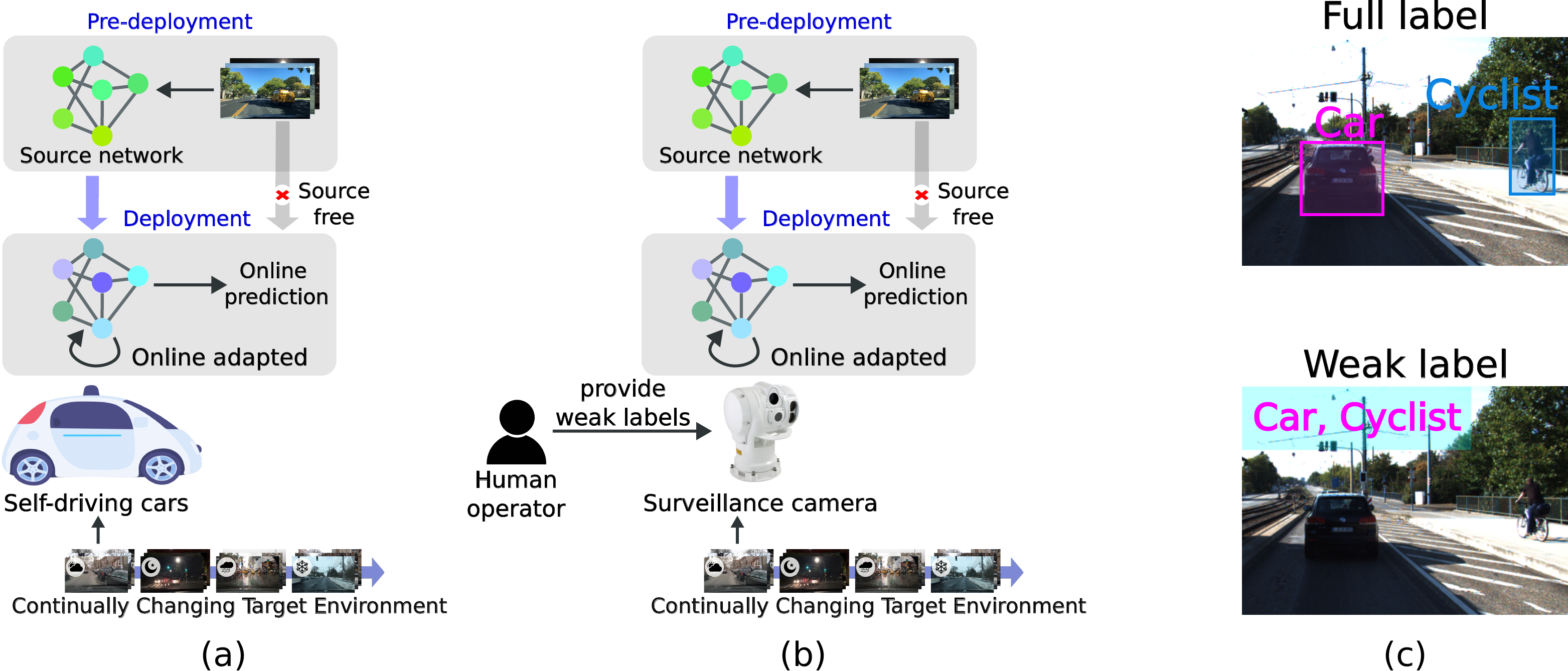}
    \caption{\textbf{(a)} Previous works have focused on developing fully autonomous solutions, primarily for self-driving vehicles~\cite{cotta, dua,tent} \textbf{(b)} Our approach, however, is proposed for visual surveillance, which are typically monitored by an operator. Therefore, our method will take advantage of the operator's involvement in the adaptation process. \textbf{(c)} The definitions of full and weak labels: A full label includes bounding boxes and object categories. A weak label only indicates which object categories are present in the image. By only requiring weak labels, our method reduces the amount of labour needed significantly.}
    \label{fig:cotta_vs_human_cotta}
\end{figure*}

However, while the training dataset is important for preparing an object detector, it may not cover all possible scenarios that the detector may encounter during its operation. This incomplete coverage is attributed to the various environmental conditions that can arise, such as different times of day, weather, and seasons. These factors cause the image appearance to differ from the training dataset, leading to a significant decline in detection accuracy. 
A solution to the problem is to continuously capture new data and adapt the system~\cite{doan2020hm, dua, cotta}. However, incorporating new data presents a substantial challenge due to the absence of labels within this new data, making the adaptation of the object detector a challenging task.

A potential solution for this issue is unsupervised domain adaptation (UDA)~\cite{chen2018domain, roychowdhury2019automatic, li2022cross}, which formulates the training dataset as the source domain and the newly acquired data as the target domain. The goal of UDA is to minimise the domain discrepancy in the feature space. Popular UDA methods include adversarial learning~\cite{chen2018domain,chen2021scale,pasqualino2021unsupervised}, optimal transport~\cite{lee2019sliced, xu2019wasserstein}, and pseudo-labelling~\cite{roychowdhury2019automatic, li2022cross, mattolin2023confmix}. However, a limitation of UDA is the offline setting: target data has to be acquired first before adapting the model for multiple epochs, whereas many practical applications necessitate domain adaptation to be done online. In addition, UDA requires complete access to the source domain, raising serious privacy and security concerns. Recent reverse engineering techniques have demonstrated that it is possible to use a limited amount of information about the data to fully recover the original data~\cite{mahendran2015understanding, dosovitskiy2016inverting, pittaluga2019revealing}. In data-driven approaches, data can be viewed as a vital asset of businesses; thus storing source data in deployed systems is indeed a hazardous undertaking.

To address the issues of UDA, test-time domain adapation (TTA) attempts to adapt the model to the target domain without the need for the source dataset~\cite{tent,cotta,dua}. Recent studies have demonstrated that TTA can be highly effective in image classification by adapting the model with pseudo-labelling and entropy minimisation~\cite{chen2022contrastive, tent}. However, TTA requires full access to the target data while in practice, the target data is usually in the form of stream, resulting in the target distribution continually evolving. To address this challenge, CoTTA~\cite{cotta} and DUA~\cite{dua} have been proposed. These methods only require an incoming target sample to adapt the model, making them suitable for online adaptation. The effectiveness of CoTTA has been demonstrated in image classification and segmentation, while DUA has been shown to be effective in object detection.

Despite their great potential, CoTTA and DUA are striving for a fully autonomous solution, which is suitable for applications such as self-driving cars. However, there are some applications, such as surveillance, which usually have a human operator overseeing the system~\cite{bloisi2016enhancing}. This raises a question of whether we should involve this operator to TTA. One benefit of human-in-the-loop TTA is to revise the pseudo-labels used for adapting the object detector. As shown in previous works~\cite{li2022cross, chen2022contrastive, litrico2023guiding}, pseudo-labelling is an effective approach for domain adaptation. However, if pseudo-labels are noisy, the object detector's error will accumulate, leading to a decline in the detection accuracy. Therefore, an on-line operator can be another reliable annotator for revising pseudo-labels. Efficient use of human contributions with minimal demands on labour cost in TTA is thus a critical objective. Our idea is illustrated in Fig.~\ref{fig:cotta_vs_human_cotta}.

\paragraph{\bf Contributions} 
This paper proposes the inclusion of humans in the test-time domain adaptation. Our method, dubbed weakly supervised test-time domain adaptation (WSTTA), uses weak labels provided by humans to guide the domain adaptation during the testing phase. As WSTTA only requires weak labels to be effective; its demand on labour cost is therefore minimal. Furthermore, WSTTA is proposed to accommodate the stream setting, where each target sample is observed only once. The experiments show that with only a few target test images, the WSTTA outperforms existing fully autonomous solutions in standard benchmarks. We hope that this encouraging result will motivate further research in WSTTA.

\section{Related work}

\begin{table*}[t]
    \centering
    \scriptsize
    \caption{A comparison of WSTTA with related approaches.}
    \begin{tabular}{lcccc}
        \toprule
        Approach & Source & Target & Streaming & Human in  \\
        & data & data & data & the loop \\
        \midrule
         Unsupervised domain adaptation~\cite{chen2018domain} & \yes & \yes & \no & \no \\
         Weakly supervised domain adaptation~\cite{inoue2018cross} & \yes & \yes & \no & Provide weak labels \\
         Test-time domain adaptation~\cite{dua} & \no & \yes & \yes & \no \\
         Active domain adaptation~\cite{su2020active} & \yes & \yes & \no & Provide full labels \\
         Source-free active domain adaptation~\cite{li2022source} & \no & \yes & \no & Provide full labels \\
         \midrule
         Weakly supervised test-time domain adaptation & \no & \yes & \yes & Provide weak labels \\
         \bottomrule
    \end{tabular}
    \label{tab:novelty}
\end{table*}

This section will review the main approaches to domain adaptation for object detection. In addition, we will
discuss the novelty of WSTTA in comparison to these approaches, which are succinctly outlined in Table~\ref{tab:novelty}.

\subsection{Unsupervised domain adaptation}
\label{sec:related_work:UDA}

Given a labelled source dataset and an unlabelled target dataset, UDA seeks to adapt an object detector to perform accurately in the target domain. There are three main techniques in UDA for object detection: adversarial learning~\cite{chen2018domain,chen2021scale,pasqualino2021unsupervised}, optimal transport~\cite{lee2019sliced, xu2019wasserstein}, and pseudo-labelling~\cite{roychowdhury2019automatic, li2022cross, mattolin2023confmix}.

Adversarial learning attempts to minimise the domain discrepancy in the feature space. To this end, domain adaptive Faster-RCNN~\cite{chen2018domain} employs gradient reversal layers~\cite{grl} in their adversarial learning framework to align the feature and instance distributions of source and target domains. This concept is further improved in~\cite{chen2021scale}, which aligns the source and target distributions across different image scales. Additionally, adversarial learning has been explored in anchor-free object detection techniques~\cite{pasqualino2021unsupervised,pasqualino2022unsupervised}. In comparison to adversarial learning which solves a minimax optimisation problem, optimal transport instead minimises Wasserstein distance between source and target domains. For instance, \cite{lee2019sliced} employs the sliced Wasserstein distance to address high-dimensional issue in the feature space and \cite{xu2019wasserstein} considers the duality of the Wasserstein distance, which can be approximated by neural networks under certain conditions. Another approach that has been shown to be effective is pseudo-labelling. To generate reliable pseudo-labels for target dataset, \cite{roychowdhury2019automatic} fuses the results of detection and tracking as well as proposes a label smoothing technique. To further improve the detection performance, some recent methods integrate pseudo-labelling to other strategies. For instance, \cite{li2022cross} incorporates pseudo-labelling to student-teacher architecture and \cite{mattolin2023confmix} combines pseudo-labelling with domain mixing techniques. 

Despite its great potential, UDA is done offline, which is not suitable for many practical applications dealing with streaming data. Furthermore, the need for access to the source data can be a shortcoming in terms of privacy and security (see Sec.~\ref{sec:intro}). To address this, our WSTTA can be performed online and is a source-free method, thus avoiding any privacy and security risks.

\subsection{Test-time domain adaptation} TTA attempts to adapt the model in an online manner without using the source data. Some interesting TTA works include Tent~\cite{tent} which proposes to update batch normalisation layers using entropy minimisation, CoTTA~\cite{cotta} which uses teacher-student architecture with pseudo-labelling to adapt the model, and DDA~\cite{gao2023back} which employs diffusion models to transform the appearance of
images from the target domain to resemble the source domain. Since these methods are only tested in image classification, DUA~\cite{dua} shows that TTA can be effectively applied to object detection by introducing a momentum decay parameter to stabilise the domain adaptation process.

As alluded, existing TTA works aim for developing fully autonomous domain adaptation techniques, which are useful to applications like self-driving cars. However, there are some other applications, such as surveillance, which usually have an operator overseeing the systems~\cite{bloisi2016enhancing}. Therefore, our WSTTA proposes to leverage this operator to generate more reliable pseudo-labels, which can be useful for TTA.

\subsection{Weakly supervised domain adaptation} 

Weakly supervised domain adaptation (WSDA) is an approach related to the concept of human-in-the-loop domain adaptation for object detection. 
This approach involves asking annotators to provide weak labels for the target dataset (see Fig.~\ref{fig:cotta_vs_human_cotta}c). Then, domain adaptation can be done using the source dataset with full labels and the target dataset with weak labels. Few approaches include combining predictions of the source pre-trained detector with weak labels to generate high-quality pseudo-labels for target images~\cite{inoue2018cross} or minimising the domain gap by using domain and weak label classifiers~\cite{xu2022h2fa}.

Through using weak labels, WSDA is shown to outperform UDA. However, WSDA also suffers drawbacks similar to those of UDA, i.e., the domain adaptation is done offline and the need to access the source data raises
privacy and security concerns (see Sec.~\ref{sec:intro}). Therefore, WSTTA is proposed to overcome these challenges.

\subsection{Active domain adaptation} \label{sec:related_work:ada}
Active domain adaptation (ADA) represents another approach aligned with the concept of human involvement in the domain adaptation. The objective of ADA is to select a subset of unlabelled target samples for manual annotation, facilitating domain adaptation. Early ADA methods~\cite{su2020active, fu2021transferable} propose measuring the domainness of each target sample using a domain discriminator in the adversarial training framework. Bi3D~\cite{yuan2023bi3d} extends this approach to 3D object detection. Recently, DiaNA~\cite{huang2023divide} proposes to evaluate domainness and uncertainty of every target sample through a unified metric.

While ADA shows promising potential, it operates within a framework similar to UDA, necessitating access to source data. This poses significant concerns about privacy and security (see Sec.~\ref{sec:intro}). In contrast, our WSTTA method is entirely source-free, mitigating any such risks associated with privacy and security.

\subsection{Source-free active domain adaptation}
Source-free active domain adaptation (SFADA)~\cite{li2022source, wang2023mhpl} attempts to perform active domain adaptation with the absence of the source data. Particularly, ELPT~\cite{li2022source} employs energy models to evaluate the free energy of each target sample. Target samples with the highest free energy levels are selected for manual annotation. MHPL~\cite{wang2023mhpl} selects active samples according to three characteristics: uncertainty, diversity, and source-dissimilarity. 

SFADA offers a solution that circumvents security and privacy concerns by eliminating the need for source data in domain adaptation. However, SFADA requires access to a pool of target data to enable the selection of a subset for manual annotation. By contrast, our WSTTA is designed for streaming data~\cite{cotta, burnafterread,streamactivelearning}, where each target sample is observed just once for model adaptation. In addition, to alleviate labour cost, we opt for weak labels provided by human---different from the target subset selection process employed in ADA and SFADA.

\section{Method}
\label{sec:method}

\begin{figure*}[t]
    \centering
    \includegraphics[width=1.0\textwidth]{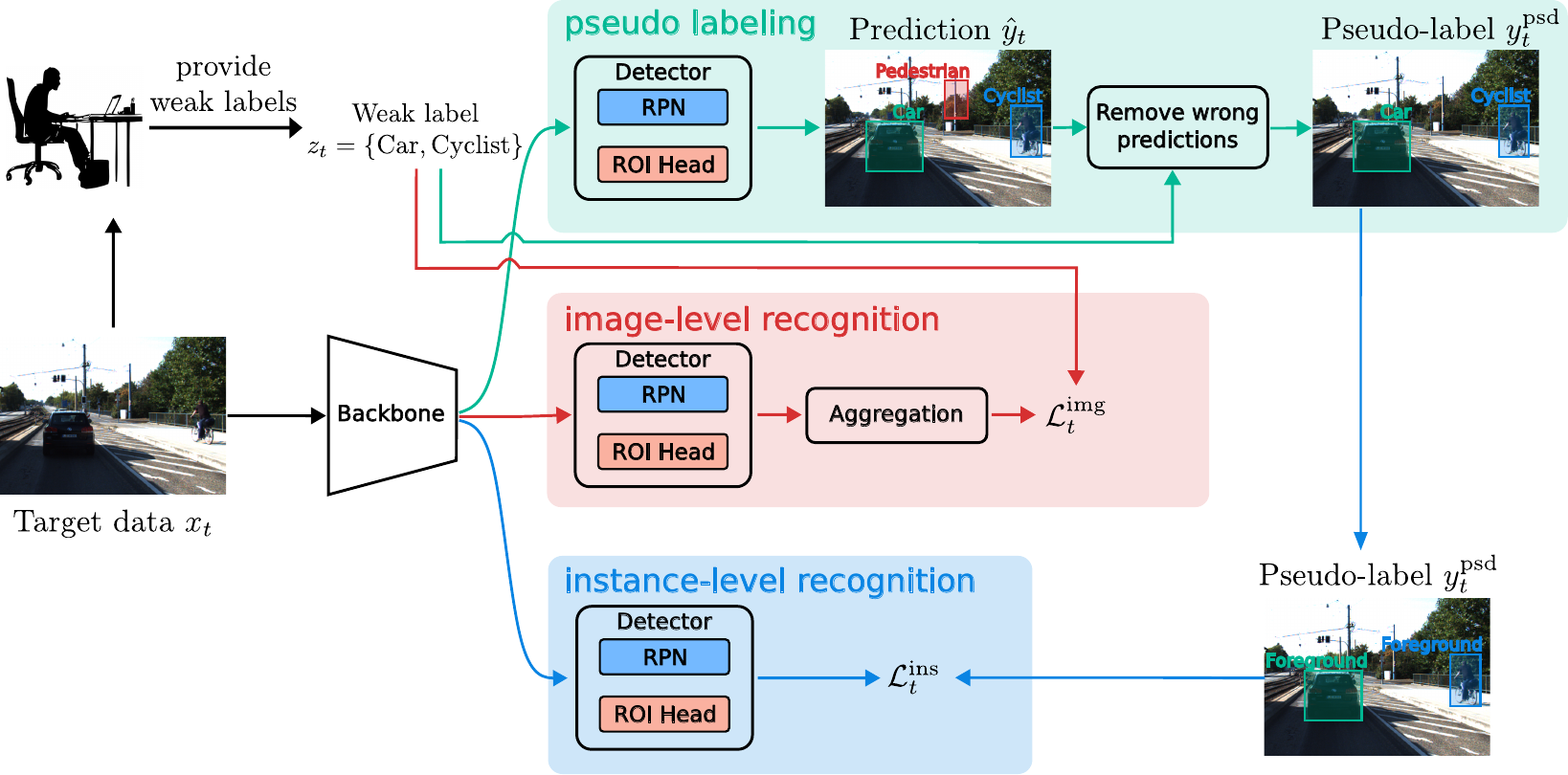}
    \caption{For an incoming target testing sample $x_t$, WSTTA initially produces a prediction $\hat{y}_t$ and the operator is required to provide a weak label $z_t$ for it. Subsequently, using the prediction $\hat{y}_t$ and weak label $z_t$, a pseudo-label $y^\text{psd}_t$ is generated. Finally, the weak label $z_t$ and pseudo-label $y^\text{psd}_t$ are used as groundtruth for image-level recognition and instance-level recognition respectively.}
    \label{fig:hg_tta}
\end{figure*}

This section will elaborate our methodology. To begin with, Sec.~\ref{sec:problem_formulation} outlines the formal problem definition. Subsequently, the WSTTA framework is presented in Sec.~\ref{sec:framework}, where we will explain the loss function for domain adaptation. Given the loss function, we will discuss how the domain adaptation can be achieved via updating batch normalisation (BN) layers in Sec.~\ref{sec:batch_norm}.

\subsection{Problem Formulation}
\label{sec:problem_formulation}

Let $f(\cdot\,;\,\theta_0)$ with parameters $\theta_0$ be an object detector that has been trained on the labelled source dataset $\left(\cX^\bS, \cY^\bS\right)$, where $\cX$ and $\cY$ are the sample space and label space. During its operation, the detector will carry out the online inference and adaptation on the unlabelled target data $\cX^\bT$. Specifically, at time step $t$, the target data $x_t \in \cX^\bT$ is given as an input to the detector $f(\cdot\,;\,\theta_t)$. Then, the detector $f(\cdot\,;\,\theta_t)$ must make an inference $\hat{y}_t = f(x_t;\theta_t)$ and adapt itself $\theta_t \rightarrow \theta_{t+1}$ for the next input $x_{t+1}$. The detector's performance is evaluated based on the predictions $\hat{y}_t$ from the online inference. 

It is noteworthy that this streaming setting closely aligns with existing literature~\cite{streamactivelearning,cotta,burnafterread}. Furthermore, it is recommended that $x_t$ should be deleted immediately after the adaptation to safeguard the privacy~\cite{burnafterread}.

The impetus for online adaptation is derived from practical scenarios in which perception systems are constantly operating in ever-evolving environments, with input coming in the form of streaming data. Moreover,
previous studies~\cite{dua,cotta,gao2023back} have mainly focused on developing fully autonomous TTA solutions for applications such as self-driving cars or autonomous robots. However, in certain applications, such as surveillance, a human operator is usually needed to supervise the system~\cite{bloisi2016enhancing}. Therefore, our idea is to involve the operator in TTA. Specifically, let $\cC$ be a set of $L$ object categories. For each target image $x_t$, the operator will provide a weak label $z_t = \{c_j\}_{j=1}^M$, where $c_j \in \cC$ is the object category present in the image and $M$ denotes the total number of object categories in the image; see Fig.~\ref{fig:cotta_vs_human_cotta}c. This weak label $z_t$ will then be leveraged to adapt the object detector's parameters $\theta_t \rightarrow \theta_{t+1}$ for the next input. As discussed in previous works~\cite{cao2021cat,vo2022active,zhong2020boosting}, providing weak labels instead of full labels will significantly reduce the amount of labour required.

\subsection{Framework}
\label{sec:framework}
The overview of WSTTA is shown in Fig.~\ref{fig:hg_tta}. Specifically, for the model $f(\cdot\,;\,\theta_t)$, WSTTA adopts a two-stage object detection architecture Faster-RCNN~\cite{fasterrcnn} that includes a backbone, a region proposal network (RPN), and a detection head (ROI Head). Our WSTTA consists of three main components: pseudo-labelling, image-level recognition, and instance-level recognition.

Using the prediction $\hat{y}_t$ and weak label $z_t$, pseudo-labelling will generate a pseudo-label $y^\text{pds}_t$. The pseudo label and weak label will be used to construct loss functions $\cL^\text{ins}_t$ in instance-level recognition and $\cL^\text{img}_t$ in image-level recognition. The final loss for domain adaptation will be
\begin{align}
    \cL_t = \cL^\text{ins}_t + \alpha.\cL^\text{img}_t
    \label{eq:final_loss}
\end{align}
This loss $\cL_t$ will be used to update Faster-RCNN's parameters $\theta_t \rightarrow \theta_{t+1}$ for the next input $x_{t+1}$ (see Sec.~\ref{sec:batch_norm}). In this section, we will outline each component: pseudo-labelling, image-level recognition, and instance-level recognition.

\paragraph{\bf Pseudo-labelling} The target image $x_t$ is initially given to the operator and the operator must provide a weak label $z_t$. Then, WSTTA makes a prediction  $\hat{y}_t = \{\hat{\bb}_i, \hat{c}_i, \hat{p}_i\}_{i=1}^N = f(x_t\,;\,\theta_t)$, where $\hat{\bb}_i \in \bbR^4$ is the predicted bounding box, $\hat{c}_i \in \cC$ is the predicted object category, $\hat{p}_i \in \bbR$ is the probability that $\hat{\bb}_i$ belongs to $\hat{c}_i$, and $N$ is the total number of predicted bounding boxes. Note that $\hat{y}_t$ is obtained after excluding overlapping boxes by non-maximum suppression for each object category.

However, the prediction $\hat{y}_t$ may contain mistakes, i.e., bounding boxes with incorrect object categories. If we use $\hat{y}_t$ as the groundtruth to adapt $\theta_t$, the errors will accumulate over time, leading to a decrease in the detector's performance. To minimise these errors, we will create a pseudo-label $y^\text{psd}_t$ by keeping bounding boxes of $\hat{y}_t$ such that their object categories are present in the weak label $z_t$ and their predicted probability is greater than $0.8$
\begin{align}
     y^\text{psd}_t = \{\hat{\bb}_i, \hat{c}_i \mid  \hat{\bb}_i, \hat{c}_i, \hat{p}_i \in \hat{y}_t \,\, \text{and} \,\, \hat{p}_i \ge 0.8 \,\, \text{and} \,\, \hat{c}_i \in z_t \}
     \label{eq:pseudo_label}
\end{align}
These pseudo-label $y^\text{psd}_t$ and weak label $z_t$ will be respectively used as groundtruth in the instance-level recognition and image-level recognition.

\paragraph{\bf Image-level recognition} This component aggregates the outputs of RPN and ROI head to obtain an image-level prediction, which is used to calculate the image-level loss $\cL^\text{img}_t$. This aggregation operation is developed based on the idea of weakly-supervised object detection~\cite{bilen2016weakly,xu2022h2fa}. Recall that $L$ is the total number of object categories, we denote $K$ as the total number of proposals, the output of RPN as $\bo \in \bbR^K$, and the output of ROI Head as $\bC \in \bbR^{K \times L}$.

Firstly, we create a matrix $\bO$ that has a same size as $\bC$
\begin{align}
    \left[\bO\right]_{k,l^\prime} = 
    \begin{cases}
    \left[\bo\right]_k \;\;\;\;\; \text{if} \;\;\;\;\; l^\prime = \underset{l}{\argmax} \, \left[\bC\right]_{k,1:L} \\
    0 \;\;\;\;\; \text{otherwise}
    \end{cases}
\end{align}
where, $\left[\cdot\right]_{k,l}$ denotes the element in the row $k^\text{th}$ and column $l^\text{th}$ of a matrix, and $\left[\cdot\right]_k$ denotes the $k^\text{th}$ element of a vector.

Next, the softmax is applied on $\bC$ and $\bO$
\begin{align}
    & \left[\sigma(\bC)\right]_{k,l} = \frac{e^{\left[\bC\right]_{k,l}}}{\sum_{l=1}^L e^{\left[\bC\right]_{k,l}}},
    \,\,\,\,\, \left[\sigma(\bO)\right]_{k,l} =  \frac{e^{\left[\bO\right]_{k,l}}}{\sum_{k=1}^K e^{\left[\bO\right]_{k,l}}}
\end{align}

Then, the image-level prediction $\hat{z}_t \in \bbR^L$ is calculated
\begin{align}
    \left[\hat{z}_t\right]_l = \sum_{k=1}^K\left[\sigma\left(\bC\right) \odot \sigma\left(\bar{\bO}\right)\right]_{k,l}
\end{align}

Finally, the image-level loss can be obtained via the standard cross-entropy function 
\begin{align}
    \cL^\text{img}_t = \texttt{cross\_entropy\_loss}\left(\hat{z}_t, \texttt{multi\_hot}\left(z_t\right)\right)
    \label{eq:img_loss}
\end{align}
where, $\texttt{multi\_hot}(\cdot)$ is a function to convert $z_t$ into a multi-hot vector of size $L$.

\paragraph{\bf Instance-level recognition}
This component will employ the pseudo-label $y^\text{psd}_t$ from Eq.~\eqref{eq:pseudo_label} as the groundtruth. The instance-level loss is formulated as follows
\begin{align}
    \cL_t^\text{ins} = \cL^\text{rpn}_\text{cls}(x_t, y^\text{psd}_t) + \cL^\text{roi}_\text{cls}(x_t, y^\text{psd}_t)
    \label{eq:ins_loss}
\end{align}
where, $\cL^\text{rpn}_\text{cls}$ and $\cL^\text{roi}_\text{cls}$ are the classification losses of RPN and ROI Head proposed in standard Faster-RCNN~\cite{fasterrcnn}. Here, the instance-level loss $\cL_t^\text{ins}$ ignores the bounding-box regression task since the pseudo-label $y^\text{psd}_t$ in Eq.~\eqref{eq:pseudo_label} indicates the confidence score of predicted bounding boxes. Thus, the parameters $\theta_t$ are adapted to enhance the classification performance of the detector.

\subsection{Domain gap minimisation via updating batch normalisation} 
\label{sec:batch_norm}
Given the final loss from Eq.~\eqref{eq:final_loss}, we need to adapt $\theta_t \rightarrow \theta_{t+1}$ for the next input $x_{t+1}$. We choose to update all BN layers in $\theta$ as this has been shown to be highly effective in recent studies~\cite{dua, tent, schneider2020improving}. The rationale of updating BN layers is to reduce the covariate shift between the source and target distributions~\cite{schneider2020improving}. If the target distribution is different from source distributions, BN's parameters estimated from the source distribution are no longer normalising the target data as expected. Therefore, it is necessary to update BN layers with the new target distribution.

Specifically, for an arbitrary BN layer of $\theta_t$, let $\mu_t$ and $\sigma_t$ be its running mean and running variance, and also let $\gamma_t$ and $\beta_t$ be its transformation parameters. We also denote $m_t$ as its momentum ($m_0$ is set to 0.1 by default). As shown in~\cite{dua}, if the momentum is gradually decayed, it will stabilise the convergence of the domain adaptation. Therefore, we initially decay the momentum
\begin{align}
    m_t = m_{t-1}.\omega + \delta
\end{align}
where, $\omega \in (0,1)$ is a predefined decay parameter and $\delta$ defines the lower bound of momentum.

Subsequently, BN's parameters will be updated as follows
\begin{align}
    & \mu_{t+1} = (1-m_t).\mu_t + m.\hat{\mu}_t, 
    && \sigma_{t+1} = (1-m_t).\sigma_t + m.\hat{\sigma}_t, \label{eq:update_mean_var} \\
    & \gamma_{t+1} = \gamma_t + \lambda \frac{\partial \cL_t}{\partial \gamma_t}, 
    && \beta_{t+1} = \beta_t + \lambda \frac{\partial \cL_t}{\partial \beta_t},
    \label{eq:update_gamma_beta}
\end{align}
where, $\lambda$ is the step size of gradient update and $\hat{\mu}$ and $\hat{\sigma}$ are mean and variance of the current input data. Note that the previous work~\cite{dua} only updates $\mu_t$ and $\sigma_t$ while our method can also update transformation parameters $\gamma_t$ and $\beta_t$, thanks to weak labels provided by the operator.

\section{Experiment}

\begin{figure*}[h]
    \centering
    \subfloat[Visible images]{
        \includegraphics[width=1.0\textwidth]{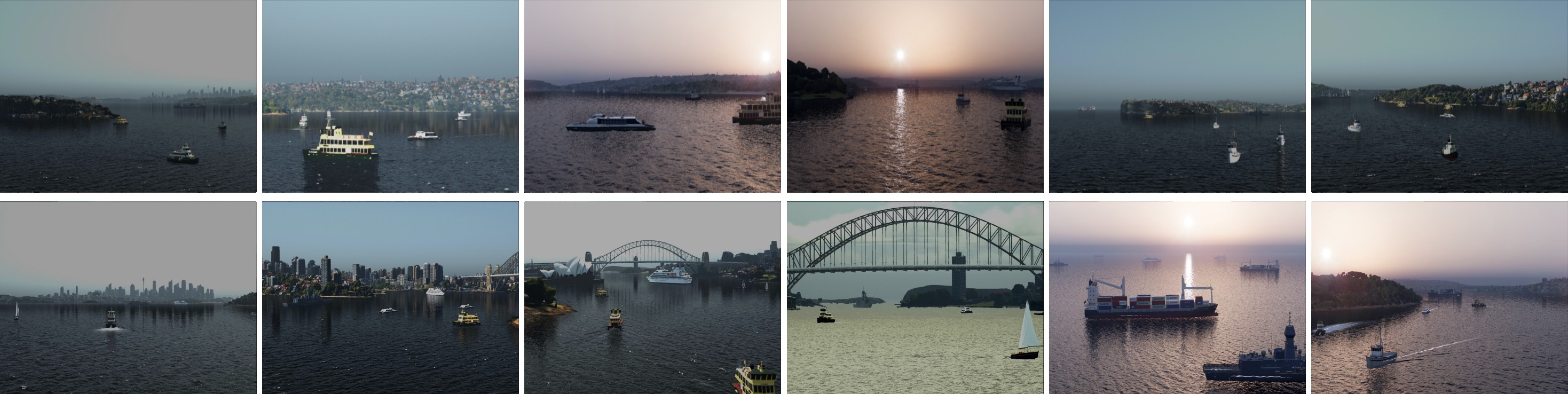}
    }
    
    \subfloat[Infrared images]{
        \includegraphics[width=1.0\textwidth]{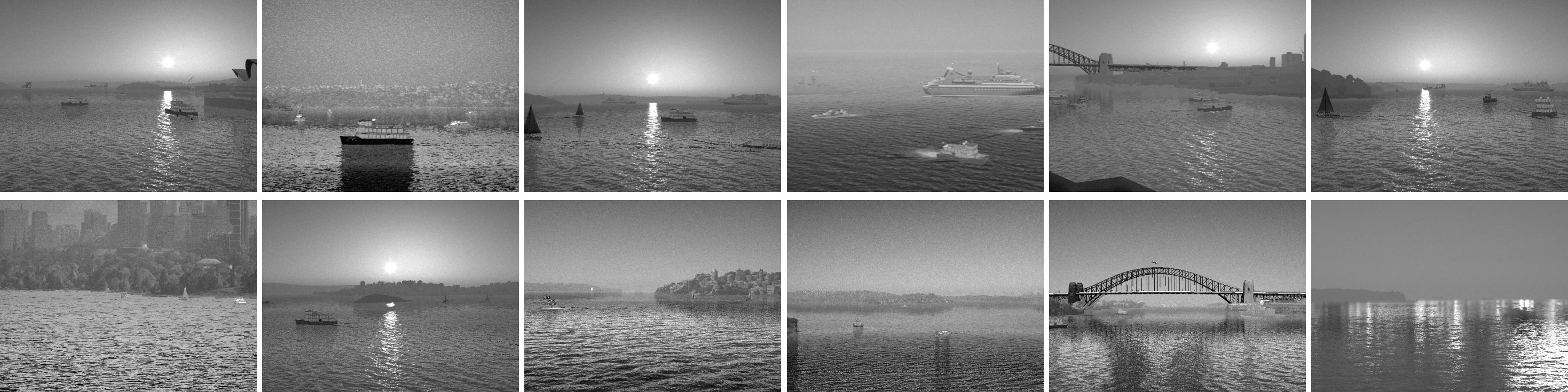}
    }
    \caption{Sample images of our MSA-SYNTH dataset, where we simulate different environmental conditions.}
    \label{fig:msa_synth}
    \vspace{-3em}
\end{figure*}

\subsection{Setup}
The following benchmarks are used in the experiment.
\begin{itemize}
    \item \textbf{KITTI $\rightarrow$ KITTI-Fog}: KITTI~\cite{kitti} is a widely-used dataset in autonomous driving. KITTI is used to pre-train the model, which are then adapted to the target KITTI-Fog with the most severe fog level 30m visibility~\cite{kittifog}. Object categories taken into account are ``Car", ``Pedestrian", and ``Cyclist". A total of 7481 images are randomly divided into 3740 for training and 3741 for testing.

    \item \textbf{Cityscapes $\rightarrow$ KITTI}: Cityscapes~\cite{cityscapes} is another popular dataset in self-driving cars. We pre-train the object detector in Cityscape, then adapt it to KITTI. Three object categories in Cityscapes are used: ``Car", ``Pedestrian", and ``Rider". Similarly, three object categories in KITTI are used: ``Car", ``Pedestrian", and ``Cyclist". A total of 3475 Cityscapes images from its training and validation sets are randomly divided into 1737 for training and 1738 for testing. Similarly, a total of 7481 KITTI images are randomly split into 3740 for training and 3741 for testing.

    \item \textbf{Visible $\rightarrow$ Infrared}~\footnote{We will release the dataset after the review; subject to all internal approvals.}: We use Unreal Engine~\footnote{\url{https://www.unrealengine.com/}} and Infinite Studio~\footnote{\url{https://infinitestudio.software/}} to generate a maritime dataset MSA-SYNTH. Three boat/vessel categories ``Fishing", ``Sailing", and ``Passenger" are simulated. We collect 8147 visible images and 8147 infrared images, which are then divided into 4243 visible images for training, 3904 visible images for testing, 4243 infrared images for training, and 3904 infrared images for testing; see Fig.~\ref{fig:msa_synth} for sample images. The model will be pre-trained in the source domain ``Visible", then adapted to the target domain ``Infrared".
\end{itemize}

\begin{wrapfigure}{r}{0.5\textwidth}
    \vspace{-2em}
    \captionof{table}{Comparing AP50 within each object categories and mAP across all categories between WSTTA and other baselines (larger is better)} 
    \scriptsize
    \centering
    \subfloat[KITTI $\rightarrow$ KITTI-Fog]{
        \begin{tabular}{ lC{1cm}C{1.4cm}C{1.3cm}C{0.8cm} }
            \toprule
            & \textbf{Car} & \textbf{Pedestrian} & \textbf{Cyclist} & \textbf{mAP} \\
            \midrule
            \texttt{Source} & 23.4 & 26.7 & 12.4 & 20.9 \\
            \texttt{BN Stats} & 41.3 & 41.4 & 20.8 & 34.5 \\
            \texttt{DUA} & 41.3 & 41.8 & 21.3 & 34.8 \\
            \texttt{WSTTA} & 44.6 & 41.9 & 23.1 & 36.5\\
            \midrule
            \texttt{Oracle} & 85.5 & 65.7 & 68.3 & 73.2\\
            \bottomrule
        \end{tabular}
    }
    \vspace{1em}
    \subfloat[Cityscapes~$\rightarrow$~KITTI]{
        \begin{tabular}{ lC{1cm}C{1.4cm}C{1.3cm}C{0.8cm} }
            \toprule
            & \textbf{Car} & \textbf{Pedestrian} & \textbf{Cyclist} & \textbf{mAP} \\
            \midrule
            \texttt{Source} & 66.9 & 46.4 & 9.0 & 40.8 \\
            \texttt{BN Stats} &  68.1 & 50.1 & 12.3 & 43.5\\
            \texttt{DUA} &  68.1 & 50.3 & 12.7 & 43.7\\
            \texttt{WSTTA} &  68.1 & 51.5 & 14.3 & 44.6\\
            \midrule
            \texttt{Oracle} & 90.4 & 70.7 & 77.2 & 79.4\\
            \bottomrule
        \end{tabular}
    }
    \vspace{1em}
    \subfloat[Visible $\rightarrow$ Infrared]{
        \begin{tabular}{ lC{1cm}C{1.4cm}C{1.3cm}C{0.8cm} }
            \toprule
            & \textbf{Fishing} & \textbf{Sailing} & \textbf{Passenger} & \textbf{mAP} \\
            \midrule
            \texttt{Source} &  42.4 & 15.6 & 33.5 & 30.5\\
            \texttt{BN Stats} & 43.8 & 14.8 & 37.0 & 31.8 \\
            \texttt{DUA} &  44.6 & 15.1 & 37.0 & 32.2 \\
            \texttt{WSTTA} & 54.7 & 21.1 & 36.2 & 37.4 \\
            \midrule
            \texttt{Oracle} & 69.9 & 39.2 & 72.8 & 60.6 \\
            \bottomrule
        \end{tabular}
    }
    
    \label{tab:benefits_human_guide}
    \vspace{-2em}
\end{wrapfigure} 
We consider following baselines
\begin{itemize}
    \item \textbf{\texttt{Source}}: The source pre-trained model is tested on the target data without any adaptation.
    \item \textbf{\texttt{BN stats}}: BN stats~\cite{schneider2020improving} adapts the source pre-trained model by updating the statistics of batch normalisation (BN) layers.
    \item \textbf{\texttt{DUA}}: DUA~\cite{dua} introduces a decay factor to update the momentum parameters of the BN layers of the source pre-trained model. 
    \item \textbf{\texttt{Oracle}}: The source pre-trained model is fine-tuned in 120k iterations on the target training set with full supervision.
\end{itemize}
where, \textbf{\texttt{BN stats}}~\cite{schneider2020improving} and \textbf{\texttt{DUA}}~\cite{dua} are fully autonomous adaptation methods which use BN update to minimise the domain gap. \textbf{\texttt{Source}} and \textbf{\texttt{Oracle}} provide the lower-bound and upper-bound performance.

To reduce labour cost, our WSTTA uses 100 target testing images for adaptation in all benchmarks, unless otherwise stated. For baselines, all target testing images are used for adaptation since they are fully autonomous adaptation techniques. Following the streaming setting~\cite{streamactivelearning,cotta,burnafterread}, each target testing image is given to each method one at a time to perform domain adaptation. 

To measure object detection performance, we present the average precision with a threshold of 50\% (AP50) for each object category and the mean average precision (mAP) across all object categories.

\subsection{Implementation}
We employ Detectron2~\cite{detectron2} for implementation. Faster-RCNN with backbone ResNet-50~\cite{fasterrcnn} is pre-trained on the source dataset with a batch size of 2. Learning rate is initially set to 0.001 for the first 30,000 iterations, then reduced to 0.0001 for the remaining 90,000 iterations. 

For WSTTA, unless stated otherwise we set $\omega = 0.99$ for KITTI~$\rightarrow$~KITTI-Fog and Cityscapes~$\rightarrow$~KITTI, and $\omega = 0.94$ for Visible~$\rightarrow$~Infrared. For remaining parameters, unless stated otherwise we set learning rate $\lambda = 0.0001$, $\delta = 0.005$, and $\alpha = 0.1$ for all benchmarks.

\subsection{Results}
\subsubsection{Benefits of human guidance in TTA}

As shown in Table~\ref{tab:benefits_human_guide}, the fully autonomous TTA methods \texttt{BN Stats} and \texttt{DUA} outperforms \texttt{Source} by 11\%-14\% mAP in all benchmarks.

When human guidance is incorporated into the TTA, \texttt{WSTTA} increases mAP by 4\% mAP in Visible~$\rightarrow$~Infrared and 1\% in KITTI~$\rightarrow$~KITTI-Fog and Cityscapes $\rightarrow$~KITTI, compared to \texttt{BN Stats} and \texttt{DUA}. The improvement is even more significant for certain object categories. For instance, \texttt{WSTTA} improves  ``Car'' in KITTI~$\rightarrow$~KITTI-Fog, ``Cyclist'' in Cityscapes~$\rightarrow$~KITTI, and ``Fishing'' in Visible~$\rightarrow$~Infrared by about 3.3\%, 2\%, and 10\% respectively, compared to \texttt{BN Stats} and \texttt{DUA}.

\begin{wrapfigure}{r}{0.5\textwidth}
    \centering
    \includegraphics[width=0.45\textwidth]{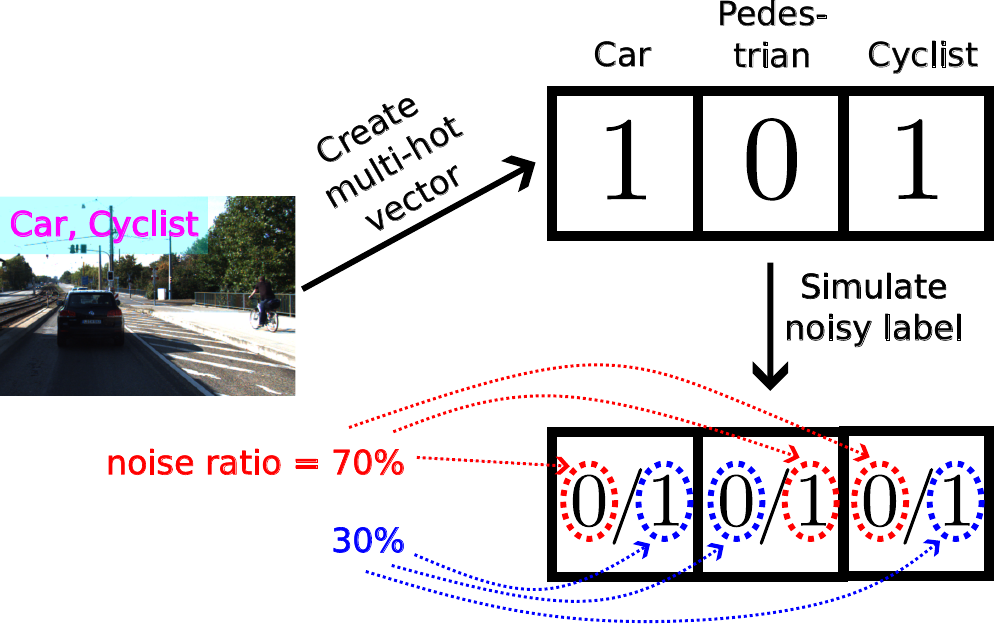}
    \caption{Illustration of how noisy weak labels are simulated. Given a weak label \{Car, Cyclist\}, a corresponding multi-hot vector is created. If the noise ratio is 70\%, the value 1 in ``Car" element will have the 70\% probability of being switched to 0, while having the 30\% probability of remaining 1. A similar operation is applied to elements ``Pedestrian" and ``Cyclist".}
    \label{fig:simulate_noise}
    \vspace{-2em}
\end{wrapfigure} 

\texttt{BN Stats} and \texttt{DUA} only updates the running mean and variance using Eq.~\eqref{eq:update_mean_var} while \texttt{WSTTA} also performs the gradient update on transformation parameters using Eq.~\eqref{eq:update_gamma_beta}. The improvements shown in Table~\ref{tab:benefits_human_guide} suggest that the loss $\cL_t$ of Eq.~\eqref{eq:final_loss} constructed by using weak labels significantly enhances the TTA's performance.

\subsubsection{Effects of noisy weak labels}

\begin{figure}
    \centering
    \subfloat[]{
        \begin{tikzpicture}[scale=0.48]
            \begin{axis}[
                xlabel=Noise percentage (\%),
                ylabel=AP50,
                ylabel near ticks,
                xmin=0, xmax=2,
                ymin=0, ymax=70,
                xtick={0,1,2},
                xticklabels={0, 50, 99},   
                ytick={0,10,...,100},
                legend style={at={(0.03,0.16)},anchor=west}]
                \addplot[smooth,mark=*,blue, line width=2pt] 
                    plot coordinates {
                        (0,44.6)
                        (1,44.0)
                        (2,38.2)
                    };
                \addlegendentry{Car}
               \addplot[smooth,color=red,mark=x, line width=2pt]
                    plot coordinates {
                        (0,41.9)
                        (1,42.3)
                        (2,41.9)
                    };
                \addlegendentry{Pedestrian}

                \addplot[smooth,color=green,mark=triangle, line width=2pt]
                    plot coordinates {
                        (0,23.1)
                        (1,22.0)
                        (2,19.8)
                    };
                \addlegendentry{Cyclist}
            \end{axis}
        \end{tikzpicture}
        \label{fig:results_noise:kitti_to_kittifog}
    }
    \subfloat[]{
        \begin{tikzpicture}[scale=0.48]
            \begin{axis}[
                xlabel=Noise percentage (\%),
                ylabel=AP50,
                ylabel near ticks,
                xmin=0, xmax=2,
                ymin=0, ymax=70,
                xtick={0,1,2},
                xticklabels={0, 50, 99},   
                ytick={0,10,...,100},
                legend style={at={(0.03,0.35)},anchor=west}]
                \addplot[smooth,mark=*,blue, line width=1pt] 
                    plot coordinates {
                        (0,68.1)
                        (1,67.7)
                        (2,62.1)
                    };
                \addlegendentry{Car}
                
                \addplot[smooth,color=red,mark=x, line width=1pt]
                    plot coordinates {
                        (0,51.5)
                        (1,50.3)
                        (2,46.5)
                    };
                \addlegendentry{Pedestrian}
        
                \addplot[smooth,color=green,mark=triangle, line width=1pt]
                    plot coordinates {
                        (0,14.3)
                        (1,13.3)
                        (2,12.2)
                    };
                \addlegendentry{Cyclist}
            \end{axis}
        \end{tikzpicture}
        \label{fig:results_noise:cityscapes_to_kitti}
    }
    \subfloat[]{
        \begin{tikzpicture}[scale=0.48]
            \begin{axis}[
                xlabel=Noise percentage (\%),
                ylabel=AP50,
                ylabel near ticks,
                xmin=0, xmax=2,
                ymin=0, ymax=70,
                xtick={0,1,2},
                xticklabels={0, 50, 99},   
                ytick={0,10,...,100},
                legend style={at={(0.03,0.16)},anchor=west}]
                \addplot[smooth,mark=*,blue, line width=2pt] 
                    plot coordinates {
                        (0,54.7)
                        (1,48.6)
                        (2,46.3)
                    };
                \addlegendentry{Fishing}
                
                \addplot[smooth,color=red,mark=x, line width=2pt]
                    plot coordinates {
                        (0,21.1)
                        (1,18.5)
                        (2,16.3)
                    };
                \addlegendentry{Sailing}
        
                \addplot[smooth,color=green,mark=triangle, line width=2pt]
                    plot coordinates {
                        (0,36.2)
                        (1,36.8)
                        (2,35.1)
                    };
                \addlegendentry{Passenger}
            \end{axis}
            \label{fig:results_noise:rgb_to_ir}
        \end{tikzpicture}
    }
    \caption{Effects of noisy weak labels to WSTTA are shown on benchmarks \textbf{(a)} KITTI~$\rightarrow$~KITTI Fog, \textbf{(b)} Cityscapes~$\rightarrow$~KITTI, and \textbf{(c)} Visible~$\rightarrow$~Infrared.}
    \label{fig:results_noise}
\end{figure}
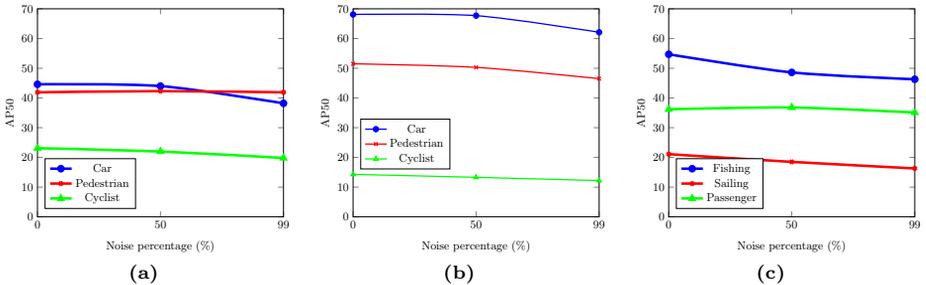

This experiment examines the possibility of humans providing incorrect weak labels. To simulate this, each element in the groundtruth multi-hot vector has a probability of being switched to an incorrect value. This probability is referred to as the noise ratio. An example of this simulation is shown Fig~\ref{fig:simulate_noise}. The results are shown in Fig.~\ref{fig:results_noise}. In general, WSTTA can be seen to be sensitive to noisy labels.

At a noise level of 50\%, the performance of WSTTA decreases by about 1\% in the ``Car'' and ``Cyclist'' categories in the KITTI~$\rightarrow$~KITTI-Fog benchmark. A similar 1\% drop is observed in the ``Pedestrian'' and ``Cyclist'' categories in the Cityscapes $\rightarrow$~KITTI benchmark. However, the most significant decrease is seen in the Visible~$\rightarrow$~Infrared benchmark, where the accuracy of the ``Fishing'' category decreases by 6\% and that of the ``Sailing'' category drops by 2.5\%.

When the noise ratio is increased to 99\%, the performance of WSTTA in KITTI~$\rightarrow$~KITTI-Fog reduces by 6\% and 3\% in the ``Car'' and ``Cyclist'' categories respectively. Similarly, there is a significant decrease of 6\% and 5\% in the ``Car'' and ``Pedestrian'' categories in Cityscapes~$\rightarrow$~KITTI. The most dramatic decline is observed in Visible~$\rightarrow$~Infrared, where the performance of WSTTA drops by more than 8\% and nearly 5\% in the ``Fishing" and ``Sailing".

The results in Fig.~\ref{fig:results_noise} shows that using only 100 images ($\sim$2.5\% of data) with noisy weak labels for domain adaptation will lead to the incorrect calculation of loss $\cL_t$ in Eq.~\eqref{eq:final_loss}, resulting in significant drop in object detection's performance. This indicates that weak labels provided by human have a substantial impact on the TTA's performance.

\begin{figure}
    \centering
    \subfloat[] {
        \begin{tikzpicture}
            \begin{axis}
            [width=6cm,
            ylabel = mAP,
            ylabel near ticks,
            ymin=20, ymax=42,
            ytick={20,25,30,35,40},
            yticklabels={20,25,30,35,40},
            xlabel = Number of samples used for adaptation,
            legend style={at={(0.27,0.17)},anchor=west, font=\tiny}, 
            legend cell align={left}]
            
            \addplot[smooth,red, line width=1pt] table[x=num_adapt,y=mAP50-Mean, col sep=comma] {data/orders.csv};
            
            \addplot [name path=upper,draw=none] table[x=num_adapt,y expr=\thisrow{mAP50-Mean}+\thisrow{mAP50-STD}, col sep=comma] {data/orders.csv};
            \addplot [name path=lower,draw=none] table[x=num_adapt,y expr=\thisrow{mAP50-Mean}-\thisrow{mAP50-STD}, col sep=comma] {data/orders.csv};
            \addplot [cyan!50!white, area legend,] fill between [of=upper and lower];
            \legend{Mean, , , Standard deviation}
            \end{axis}
        \end{tikzpicture}
        \label{fig:sample_order}
    }
    \subfloat[]{
        \begin{tikzpicture}
            \begin{axis}
            [width=6cm,
            ylabel = mAP,
            ylabel near ticks,
            ymin=20, ymax=42,
            ytick={20,25,30,35,40},
            yticklabels={20,25,30,35,40},
            xlabel = Number of samples used for adaptation,
            legend style={at={(0.55,0.31)},anchor=west,font=\tiny}, 
            legend cell align={left}]
            
            \addplot[smooth,red, line width=1.5pt] table[x=num_adapt,y=mAP50, col sep=comma] {data/1.00.csv};
            \addlegendentry{$\omega = 1.0$}
            
            \addplot[smooth,blue, line width=1.5pt] table[x=num_adapt,y=mAP50, col sep=comma] {data/0.99.csv};
            \addlegendentry{$\omega = 0.99$}
    
            \addplot[smooth,green, line width=1.5pt] table[x=num_adapt,y=mAP50, col sep=comma] {data/0.97.csv};
            \addlegendentry{$\omega = 0.97$}

            \addplot[smooth,cyan, line width=1.5pt] table[x=num_adapt,y=mAP50, col sep=comma] {data/0.95.csv};
            \addlegendentry{$\omega = 0.95$}

            \addplot[smooth,black, line width=1.5pt] table[x=num_adapt,y=mAP50, col sep=comma] {data/0.93.csv};
            \addlegendentry{$\omega = 0.93$}
    
            \addplot[smooth,magenta, line width=1.5pt] table[x=num_adapt,y=mAP50, col sep=comma] {data/0.91.csv};
            \addlegendentry{$\omega = 0.91$}
            \end{axis}
        \end{tikzpicture}
        \label{fig:decay_factors}
    }
    \caption{\textbf{(a)} WSTTA results on KITTI~$\rightarrow$~KITTI-Fog for 30 independent runs. For each run, the order of KITTI-Fog testing samples is randomly shuffled. \textbf{(b)} WSTTA results on KITTI~$\rightarrow$~KITTI-Fog for different decay factors.}
\end{figure}
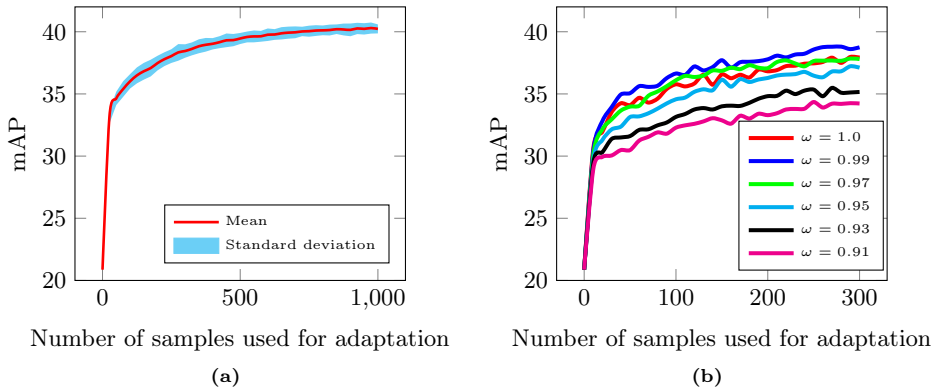

\paragraph{\bf Effects of sample orders}
This experiment investigates the effects of sample orders. The result on KITTI~$\rightarrow$~KITTI-Fog in shown in Fig.~\ref{fig:sample_order}, where we conduct 30 independent runs and calculate the mean and standard deviation of mAP. For each run, the order of samples is randomly shuffled. In the first few samples used for adaptation, the standard deviation is large. For example, with 100 samples, we obtain mAP of 36.0~$\pm$~0.6. When the adaptation samples increase, the mAP continues to improve as well as the standard deviation decreases. For instance, the mAP achieves 39.8~$\pm$~0.5 at 600 samples and 40.2~$\pm$~0.4 at 1000 samples. However, the mAP saturates at 40.0 after 700 samples. 

The small values in standard deviation indicate that sample order does not significantly affect \texttt{WSTTA}. Additionally, using more samples for domain adaptation significantly improves object detection performance but also increases labour costs. However, the improvement saturates after a certain number of samples, likely because the domain gap has been minimised at the batch normalization layers.

\paragraph{\bf Effects of decay factors}
We investigate the effect of the decay factor $\omega$ on the performance of WSTTA on the KITTI~$\rightarrow$~KITTI-Fog benchmark. Fig.~\ref{fig:decay_factors} shows the results when different decay factors $\omega$ from 1.0 (no decay) to 0.91 are applied. We observe that when $\omega$ is set to 0.99 or 0.97, the mAP is better than that of $\omega=1.0$, indicating that decaying the momentum accelerates the convergence of domain adaptation. However, if the momentum decays too quickly (i.e. $\omega < 0.97$), the detection accuracy decreases.

As shown in~\cite{dua}, decay factor is necessary to stabilise the domain adaptation. However, if it is set too small or too large, the minimisation process of the domain gap will take longer time to converge to local minima, leading smaller improvements in object detection performance. This suggests that tuning the decay factor is essential for achieving satisfactory domain adaptation performance.

\section{Conclusion and future works}

\subsubsection{Conclusion} This paper presents a method involving a human operator in TTA. The algorithm only requires the operator to provide weak labels for images, which are then used to guide the adaptation process. The experiments show that the proposed method outperforms existing autonomous test-time adaptation solutions, demonstrating great potential of human guidance for TTA.

\paragraph{\bf Future works} A promising future work is to examine the question ``when to adapt?" in the WSTTA framework. This aspect has recently captured the attention of the community~\cite{doan2024assessing,yoo2023and}. Additionally, investigating various approaches to minimise the domain gap in TTA settings, such as optimal transport~\cite{courty2017joint} or maximum mean discrepancy (MMD)~\cite{yan2017mind}, holds great potential. Recent work has demonstrated that simply minimising the distances of means and variances between source and target domains can be an effective strategy in TTA~\cite{mirza2023actmad}. Another interesting direction is to develop more effective interfaces for human interaction with AI models during TTA, as well as establish appropriate metrics to measure the cognitive workload of humans during this process. 

\bibliographystyle{splncs04}
\bibliography{main}
\end{document}